\documentclass{article}

\PassOptionsToPackage{numbers, compress}{natbib}

\usepackage[preprint]{neurips_2024}

\usepackage[utf8]{inputenc} 
\usepackage[T1]{fontenc}    
\usepackage{hyperref}       
\usepackage{url}            
\usepackage{booktabs}       
\usepackage{amsfonts}       
\usepackage{nicefrac}       
\usepackage{microtype}      
\usepackage{xcolor}         
\usepackage{amsmath}
\usepackage{multirow}
\usepackage{graphicx}

\usepackage{amsthm}
\theoremstyle{plain}
\newtheorem{theorem}{Theorem}[section]

\theoremstyle{definition}
\newtheorem{definition}[theorem]{Definition}

\theoremstyle{remark}

\title{On Lexical Invariance on Multisets and Graphs}

\author{%
  Muhan Zhang\thanks{Work done partially as a practice of my research skills; not intended for publication.} \\
  Institute for Artificial Intelligence\\
  Peking University\\
  \texttt{muhan@pku.edu.cn} \\
}

\begin{document}

\maketitle

\begin{abstract}
  In this draft, we study a novel problem, called lexical invariance, using the medium of multisets and graphs. Traditionally in the NLP domain, lexical invariance indicates that the semantic meaning of a sentence should remain unchanged regardless of the specific lexical or word-based representation of the input. For example, ``The movie was extremely entertaining'' would have the same meaning as ``The film was very enjoyable''. In this paper, we study a more challenging setting, where the output of a function is invariant to any injective transformation applied to the input lexical space. For example, multiset $\{1,2,3,2\}$ is equivalent to multiset $\{a,b,c,b\}$ if we specify an injective transformation that maps $1$ to $a$, $2$ to $b$ and $3$ to $c$. We study the sufficient and necessary conditions for a most expressive lexical invariant (and permutation invariant) function on multisets and graphs, and proves that for multisets, the function must have a form that only takes the multiset of counts of the unique elements in the original multiset as input. For example, a most expressive lexical invariant function on $\{a,b,c,b\}$ must have a form that only operates on $\{1,1,2\}$ (meaning that there are 1, 1, 2 unique elements corresponding to $a,c,b$). For graphs, we prove that a most expressive lexical invariant and permutation invariant function must have a form that only takes the adjacency matrix and a \textit{difference matrix} as input, where the $(i,j)$th element of the difference matrix is 1 if node $i$ and node $j$ have the same feature and 0 otherwise. We perform synthetic experiments on TU datasets to verify our theorems.
\end{abstract}

\section{Introduction}
Lexical invariance over multiset and graph data has certain real applications, especially when some features (IDs, names, addresses, etc.) need to be anonymized. This process is usually done by applying an unknown hashing function to the real features of the nodes of the original graph/multiset. The resulting string has no semantic meaning any more, with the only useful information being the equality/inequality between any two strings. Despite that a single hashing function is usually applied to all nodes of the same multiset/graph, the hashing function can be different or can have randomness across different multisets/graphs, due to different preprocessing procedures or is intended to further reduce the risk of re-identification. In such settings, how to design an expressive enough model that gives consistent results no matter how input features are transformed becomes important, which leads to the lexical invariance problem we study in this paper.

\section{Lexical Invariance Theory of Multisets}
We first formally define lexical invariant functions and most expressive lexical invariant functions on multisets. We consider multiset because set is a special case of multiset (where all elements appear only once) that all the theorems will apply too.
\begin{definition} \textbf{(lexical invariant multiset function)}~
Consider a multiset function $f: \mathcal{S} \mapsto \Sigma^*$, where $S$ is the space of multisets whose elements are in $\Sigma^*$, and $\Sigma^*$ is the space of all possible strings over an alphabet $\Sigma$. Suppose $\sigma: \Sigma^* \mapsto \Sigma^*$ is an injective hashing function. For simplicity, we also define $\sigma$ over a multiset to be applying $\sigma$ to every element of the multiset. Then, we say $f$ is \textit{lexical invariant} if $\forall x_1, x_2 \in \mathcal{S}$, $[\exists \sigma: x_1 = \sigma(x_2)] \Rightarrow [f(x_1) = f(x_2)]$. We say $f$ is a \textit{most expressive lexical invariant} function if $\forall x_1, x_2 \in \mathcal{S}$, $[\exists \sigma: x_1 = \sigma(x_2)] \Leftrightarrow [f(x_1) = f(x_2)]$.
\end{definition}
In other words, lexical invariant multiset functions encode multisets invariant to how the elements are hashed. However, a trivial lexical invariant function is to map all multisets to 1, or map a multiset to its cardinality, which lose great information. Therefore, what's more interesting to study is the \textbf{most expressive lexical invariant function}, which encodes multisets injectively up to hashing $\sigma$. In this way, all the information is kept in the encoding.

Now we give our first theorem characterizing the necessary and sufficient conditions for a most expressive lexical invariant multiset function.

\begin{theorem}\label{thm:set}
  $f: \mathcal{S} \mapsto \Sigma^*$ is a most expressive lexical invariant function if and only if $\forall x \in \mathcal{S}$, $f(x)$ can be written as $f'(c)$, where $c$ is the multiset of counts of unique strings in the multiset $x$, and $f'$ is an injective multiset function.
\end{theorem}
An example is given in the abstract, where both $\{1,2,3,2\}$ and $\{a,b,c,b\}$ can be transformed into the same multiset of unique counts $\{1,1,2\}$, after which we can apply a multiset function on $\{1,1,2\}$ to encode them equivalently. Now we give the formal proof in the following.
\begin{proof}
We first prove the ``if'' direction. We need to prove that if $f(x)$ can be written as $f'(c)$, then $f$ is a most expressive lexical invariant function, i.e., $[\exists \sigma: x_1 = \sigma(x_2)] \Leftrightarrow [f(x_1) = f(x_2)]$. The ``$\Rightarrow$'' is easy to prove by noticing that $c_1$ and $c_2$, the multisets of unique counts of $x_1$ and $x_2$, are the same. Thus, $f(x_1) = f'(c_1) = f'(c_2) = f(x_2)$. Now we prove the ``$\Leftarrow$'' part. We have $f'(c_1) = f'(c_2)$. Since $f'$ is injective, we know that $c_1 = c_2$. W.L.O.G., let $c_1=c_2=c=\{t_1,t_2,\ldots,t_k\}$, where $t_i$ is the times of appearances of some string in the original multiset $x_1$ (or $x_2$), and we assume there are $k$ unique strings in the original multiset. Then, $x_1$ must be able to be written in the following form: 
\[x_1=\{ \underbrace{s^{(1)}_1,\ldots,s^{(1)}_1}_{t_1\text{ times}}, \underbrace{s^{(1)}_2,\ldots,s^{(1)}_2}_{t_2\text{ times}}, \ldots,
\underbrace{s^{(1)}_k,\ldots,s^{(1)}_k}_{t_k\text{ times}}
\},\]
where $s^{(1)}_i$ is some string, and $s^{(1)}_i \neq s^{(1)}_j, \forall i\neq j$. Similarly, we can write $x_2$ as:
\[x_2=\{ \underbrace{s^{(2)}_1,\ldots,s^{(2)}_1}_{t_1\text{ times}}, \underbrace{s^{(2)}_2,\ldots,s^{(2)}_2}_{t_2\text{ times}}, \ldots,
\underbrace{s^{(2)}_k,\ldots,s^{(2)}_k}_{t_k\text{ times}}
\}.\]
With the above forms, we construct a hashing function $\sigma$ by modifying the identity function. Namely, we let $\sigma(s^{(2)}_i) = s^{(1)}_i, \forall i\in [k]$. However, still keeping $\sigma(s^{(1)}_i) = s^{(1)}_i$ will make $\sigma$ no longer injective. Fortunately, there are remaining ``empty'' slots, i.e., those $s^{(2)}_i$ strings. No matter how many slots $\sigma(s^{(2)}_i)$ occupy, there will be an equal number of slots becoming available. Take $\{a,a,b\}$ and $\{b,b,c\}$ as an example. We can construct $\sigma$ such that $\sigma(a)=b, \sigma(b)=c, \sigma(c)=a$. This way, we can always construct an injective function $\sigma$ such that $\sigma(x_2) = x_1$, proving the first direction.

Now we prove the second direction, the ``only if'' part. We need to show that every most expressive lexical invariant multiset function $f(x)$ can be written as $f'(c)$. We still write $x$ as follows:
\[x=\{ \underbrace{s_1,\ldots,s_1}_{t_1\text{ times}}, \underbrace{s_2,\ldots,s_2}_{t_2\text{ times}}, \ldots,
\underbrace{s_k,\ldots,s_k}_{t_k\text{ times}}
\},~~\text{where}~t_1\leq t_2 \leq \ldots \leq t_k.\]
Since $f$ is a most expressive lexical invariant function, there always exists an $x'$ with the form:
\[x'=\{ \underbrace{1,\ldots,1}_{t_1\text{ times}}, \underbrace{2,\ldots,2}_{t_2\text{ times}}, \ldots,
\underbrace{k,\ldots,k}_{t_k\text{ times}}
\}, ~~\text{where}~t_1\leq t_2 \leq \ldots \leq t_k\]
such that $f(x) = f(x')$, and we call $x'$ the canonical form of $x$. Then we will construct an $f'$ over the multiset of unique counts $\{t_1,t_2,\ldots,t_k\}$ and show that it gives the same output as $f(x)$. This $f'(\{t_1,t_2,\ldots,t_k\})$ first sorts the elements in the multiset in an ascending order to transform the multiset into an ordered list $[t_1,t_2,\ldots,t_k]$. Then it applies another injective function $f''$ over the list, so $f'(\{t_1,t_2,\ldots,t_k\}) = f''([t_1,t_2,\ldots,t_k])$. The list function $f''$ first injectively maps the list $[t_1,t_2,\ldots,t_k]$ into another list as follows:
\[
[\underbrace{1,\ldots,1}_{t_1\text{ times}}, \underbrace{2,\ldots,2}_{t_2\text{ times}}, \ldots,
\underbrace{k,\ldots,k}_{t_k\text{ times}}].
\]
The mapping is injective since the new list uniquely determines the original list. Finally, it uses another injective list function $f'''$ over the new list above, so $f''([t_1,t_2,\ldots,t_k]) = f'''([\underbrace{1,\ldots,1}_{t_1\text{ times}}, \underbrace{2,\ldots,2}_{t_2\text{ times}}, \ldots,
\underbrace{k,\ldots,k}_{t_k\text{ times}}])$. Since the new list has a one-to-one mapping to the multiset $\{\underbrace{1,\ldots,1}_{t_1\text{ times}}, \underbrace{2,\ldots,2}_{t_2\text{ times}}, \ldots,
\underbrace{k,\ldots,k}_{t_k\text{ times}}\}$ (the multiset uniquely determines the list), the original $f$ over this canonical-form multiset can be equivalently expressed by $f'''$ over the list. Overall, we have:
\begin{align}
f(\{\underbrace{1,\ldots,1}_{t_1\text{ times}}, \underbrace{2,\ldots,2}_{t_2\text{ times}}, \ldots,
\underbrace{k,\ldots,k}_{t_k\text{ times}}\}) 
&= f'''([\underbrace{1,\ldots,1}_{t_1\text{ times}}, \underbrace{2,\ldots,2}_{t_2\text{ times}}, \ldots,
\underbrace{k,\ldots,k}_{t_k\text{ times}}])\\
&= f''([t_1,t_2,\ldots,t_k])\\
&= f'(\{t_1,t_2,\ldots,t_k\}),
\end{align}
which proves that any $f(x)$ can be written into $f'(c)$.

\end{proof}

\section{Lexical Invariance Theory of Graphs}
We next extend the theory to graph data, where additional pairwise adjacency information is presented. Note that natural language or sequential data (e.g., list, time series) are special types of graphs that our theory apply too.
\begin{definition}
\textbf{(lexical invariant graph function)}~
Let $\mathcal{G} = \{(A, X)\}$ be the space of graphs, where $A\in \{0,1\}^{n\times n}$ is the adjacency matrix with $n$ being the node number, and $X$ is the list of $n$ node features whose elements are in $\Sigma^*$ with $\Sigma^*$ being the space of all possible strings over an alphabet $\Sigma$.
Consider a graph function $f: \mathcal{G} \mapsto \Sigma^*$. Suppose $\sigma: \Sigma^* \mapsto \Sigma^*$ is an injective hashing function, and for simplicity we also define $\sigma$ over $X$ to be applying $\sigma$ element-wise to $X$. Suppose $\pi$ is a permutation function that permutes the rows and columns of $A$ simultaneously or permutes the elements of $X$. Then, we say $f$ is \textit{permutation invariant and lexical invariant} if $\forall (A_1,X_1), (A_2,X_2) \in \mathcal{G}$, $[\exists \sigma, \pi: A_1 = \pi(A_2), X_1 = \pi(\sigma(X_2))] \Rightarrow [f(A_1,X_1) = f(A_2,X_2)]$. We say $f$ is a \textit{most expressive permutation invariant and lexical invariant} function if $\forall (A_1,X_1), (A_2,X_2) \in \mathcal{G}$, $[\exists \sigma, \pi: A_1 = \pi(A_2), X_1 = \pi(\sigma(X_2))] \Leftrightarrow [f(A_1,X_1) = f(A_2,X_2)]$.
\end{definition}

Now we present our main theorem on the necessary and sufficient conditions of a most expressive lexical invariant graph function.
\begin{theorem}\label{thm:graph}
    $f: \mathcal{G} \mapsto \Sigma^*$ is a most expressive permutation invariant and lexical invariant graph function if and only if $\forall (A,X) \in \mathcal{G}$, $f(A,X)$ can be written as $f'(A,D)$, where $D$ is the difference matrix of the same shape as $A$ whose $(i,j)$th entry is 1 if $X[i]=X[j]$ and 0 otherwise, and $f'$ is an injective function up to permutation of $A$ and $D$ (i.e., $f'(A_1,D_1)=f'(A_2,D_2) \Leftrightarrow \exists \pi: A_1=\pi(A_2), D_1=\pi(D_2)$).
\end{theorem}
Intuitively, as $X$ can be arbitrarily hashed, its concrete content should not matter, while the only useful information becomes whether two elements in $X$ are equal. This information is recorded in the newly constructed \textit{difference matrix} $D$ with the same node ordering as $A$ and $X$, so that a permutation invariant and lexical invariant function $f(A,X)$ can be transformed into a permutation invariant function $f'(A,D)$. Next we formally prove the theorem.
\begin{proof}
We first prove the ``if'' direction. We need to prove that if $f(A,X)$ can be written as an injective function $f'(A,D)$ up to permutation, then $f$ is most expressive permutation invariant and lexical invariant, i.e., $[\exists \sigma, \pi: A_1 = \pi(A_2), X_1 = \pi(\sigma(X_2))] \Leftrightarrow [f(A_1,X_1) = f(A_2,X_2)]$. To prove the ``$\Leftarrow$'' part, we notice that $f(A_1,X_1) = f(A_2,X_2) \Rightarrow \exists \pi: A_1=\pi(A_2), D_1=\pi(D_2)$. Then, $\forall i,j\in [n], [D_1]_{\pi(i),\pi(j)} = [D_2]_{i,j}$, where $\pi(i)$ outputs the new index of $i$ after applying the permutation $\pi$. As $D_1, D_2$ are the difference matrices of $X_1, X_2$,  $[D_1]_{\pi(i),\pi(j)} = [D_2]_{i,j}$ means that $[X_1]_{\pi(i)} = [X_1]_{\pi(j)} \Leftrightarrow [X_2]_{i} = [X_2]_{j}$. W.L.O.G., we write $X_1$ as:
\[X_1=[ \underbrace{s^{(1)}_1,\ldots,s^{(1)}_1}_{t_1\text{ times}}, \underbrace{s^{(1)}_2,\ldots,s^{(1)}_2}_{t_2\text{ times}}, \ldots,
\underbrace{s^{(1)}_k,\ldots,s^{(1)}_k}_{t_k\text{ times}}
],\]
with $\sum_{i\in [k]} t_i = n$. We also write $\pi(X_2)$ as:
\[\pi(X_2)=[ \underbrace{s^{(2)}_1,\ldots,s^{(2)}_1}_{t_1\text{ times}}, \underbrace{s^{(2)}_2,\ldots,s^{(2)}_2}_{t_2\text{ times}}, \ldots,
\underbrace{s^{(2)}_k,\ldots,s^{(2)}_k}_{t_k\text{ times}}
].\]
If $X_1$ and $\pi(X_2)$ are not sorted as above, there always exists a permutation $\pi'$ such that $\pi'(X_1)$ and $\pi'(\pi(X_2))$ have the above forms. Similar to the proof of Theorem~\ref{thm:set}, we construct a hashing function $\sigma$ by modifying the identity function, which satisfies $\sigma(s^{(2)}_i) = s^{(1)}_i, \forall i\in [k]$. With this hashing, we have $\pi'(X_1) = \pi'(\pi(\sigma(X_2)))$, which means $X_1 = \pi(\sigma(X_2))$. We also have $A_1 = \pi(A_2)$. Therefore, we have derived that $\exists \sigma, \pi: A_1 = \pi(A_2), X_1 = \pi(\sigma(X_2))$, proving the ``$\Leftarrow$'' part.

Still under the ``if'' direction, we prove the ``$\Rightarrow$'' part. If $\exists \sigma, \pi: A_1 = \pi(A_2), X_1 = \pi(\sigma(X_2))$, then we have $D_1 = \pi(D_2)$. This is because $[D_1]_{\pi(i),\pi(j)} = [D_2]_{i,j}$ if and only if  $[X_1]_{\pi(i)}=[X_1]_{\pi(j)} \Leftrightarrow  [X_2]_i=[X_2]_j$. Then with $D_1 = \pi(D_2)$, we can derive that $f'(A_1,D_1) = f'(A_2,D_2)$ as $f'$ is permutation invariant. We then have $f(A_1,X_1)=f'(A_1,D_1)=f'(A_2,D_2)=f(A_2,X_2)$, indicating that $f(A_1,X_1) = f(A_2,X_2)$ thus concluding the ``if'' direction.

Secondly, we prove the ``only if'' direction. We need to prove that every most expressive permutation invariant and lexical invariant function $f(A,X)$ can be written as $f'(A,D)$. With the permutation invariance property of $f$, we know $\exists \pi$ such that $f(A,X) = f(\pi(A),\pi(X))$, where $\pi(X)$ sorts $X$ to the following form:
\[
\pi(X)=[ \underbrace{s_1,\ldots,s_1}_{t_1\text{ times}}, \underbrace{s_2,\ldots,s_2}_{t_2\text{ times}}, \ldots,
\underbrace{s_k,\ldots,s_k}_{t_k\text{ times}}
].
\]
W.L.O.G., we assume $t_1\leq t_2 \leq \ldots \leq t_k$. Since $f$ is also lexical invariant, we can construct a hashing function $\sigma$ such that $\pi(\sigma(X)) = [ \underbrace{1,\ldots,1}_{t_1\text{ times}}, \underbrace{2,\ldots,2}_{t_2\text{ times}}, \ldots,
\underbrace{k,\ldots,k}_{t_k\text{ times}}
]$ (the canonical form of $X$), while $f(A,X) = f(\pi(A), \pi(\sigma(X)))$.

Define $D'$ as the difference matrix of $\pi(\sigma(X)$, which is a block-diagonal matrix as follows:
\[
D' = \begin{pmatrix}
    \begin{array}{c|c|c}
        \underbrace{1}_{t_1 \times t_1} & 0 & \cdots \\
        \hline
        0 & \underbrace{1}_{t_2 \times t_2} & \cdots \\
        \hline
        \vdots & \vdots & \ddots 
    \end{array}
\end{pmatrix}
\]
Since $D'$ uniquely determines $\pi(\sigma(X))$, we will construct an injective up to permutation function $f'(\pi(A),D')$ that always gives the same output as $f(A,X)$. This $f'$ first reconstructs $\pi(\sigma(X))$ from $D'$, and then uses an injective up to permutation function $f''$ over $(\pi(A),\pi(\sigma(X)))$, so $f'(\pi(A),D') = f''(\pi(A),\pi(\sigma(X)))$. Finally, we notice that the original $f$ over $(\pi(A), \pi(\sigma(X)))$ can be equivalently expressed by $f''(\pi(A),\pi(\sigma(X)))$. This does not mean $f = f''$, as $f$ is additionally lexical invariant. Instead, it only means that under the canonical form, $f''$ has the same output as $f$. Overall, we have:
\begin{align}
    f(A,X) &= f(\pi(A), \pi(\sigma(X))) \\
    &= f''(\pi(A),\pi(\sigma(X))) \\
    &= f'(\pi(A),D') \\
    &= f'(\pi(A),\pi(D)) \\
    &= f'(A,D).
\end{align}
The second to last step is due to $D' = \pi(D)$ (because $[D']_{\pi(i),\pi(j)}=1 \Leftrightarrow [D]_{i,j}=1$). The last step is due to the permutation invariance property of $f'$. The above equations indicate that every most expressive permutation invariant and lexical invariant function $f(A,X)$ can be expressed by an injective up to permutation function $f'(A,D)$, which concludes the proof.


\end{proof}

\section{Experiments}
We conduct synthetic experiments to verify our theory. To simulate a lexical invariant setting, we modify the node features of TU datasets by applying the sha256 hashing function to the concatenation of the original node feature and a random vector associated with each graph. For example, suppose the original feature vector of node $i$ in graph $j$ is $x_i$, and there is a pre-sampled random vector $r_j$ associated with graph $j$. Then the new feature of node $i$ will be $\texttt{sha256}(\text{concat}(x_i,r_j))$.
This ensures that different graphs have different hashing schemes, and prevents models from memorizing the same node features across different graphs. Our task is to train graph classification models under these hashed node features. Following the setting of pytorch geometric's graph classification example, we train each model for 100 epochs with batch size 128 and learning rate 0.01, and report the final loss, training accuracy and test accuracy. The code is available at \url{https://github.com/GraphPKU/LexiInv}.

We use four baseline models: DeepSet, DeepCount, GIN, and DiffGIN. DeepSet does not respect the lexical invariance, and directly applies a deep set~\cite{zaheer2017deep} model on the multiset of hashed node features of each graph. DeepCount respects the lexical invariance, and applies a deep set model to the multiset of unique counts of the hashed node features of each graph, implementing a model in Theorem~\ref{thm:set}. GIN does not respect the lexical invariance, and directly applies GIN~\cite{xu2018powerful} to each graph with hashed node features. Lastly, DiffGIN is our proposed model, which includes two GIN models operating on adjacency matrix and difference matrix separately. DiffGIN uses all-one node features in the first graph convolution layer. In each layer, we apply the two GIN models' corresponding layers with adjacency matrix and difference matrix as edges, respectively. The resulting node embeddings are summed and passed to the next layer. After all layers, a global sum pooling is used to get the graph embedding. This DiffGIN aims to implement a model in Theorem~\ref{thm:graph}, yet is not guaranteed to achieve the maximum expressivity due to the limited expressive power of GIN.

Our results are presented in Table~\ref{tab:exp}. From all the comparisons, we can clearly observe that models directly operating on the hashed features have much lower loss and higher training accuracy, but models respecting lexical invariance demonstrate less overfitting and higher testing accuracy. The experimental results \textbf{match with our theory very well}.

\begin{table}[]
\label{tab:exp}
\caption{Experimental results on three TU datasets. Units for train/test are accuracy.}
\begin{tabular}{@{}cccc|cc@{}}
\toprule
                          &       & DeepSet           & DeepCount         & GIN               & DiffGIN           \\ \midrule
\multirow{3}{*}{MUTAG}    & loss  & \textbf{0.2611}$\pm$0.0606 & 0.3833$\pm$0.0092 & \textbf{0.1258}$\pm$0.0530 & 0.1795$\pm$0.0371 \\
                          & train & \textbf{0.8615}$\pm$0.0693 & 0.8296$\pm$0.0087 & \textbf{0.9467}$\pm$0.0345 & 0.8911$\pm$0.0352 \\
                          & test  & 0.5789$\pm$0.0881 & \textbf{0.8842}$\pm$0.0394 & 0.7158$\pm$0.0632 & \textbf{0.8316}$\pm$0.0614 \\\midrule
\multirow{3}{*}{PROTEINS} & loss  & \textbf{0.3969}$\pm$0.0956 & 0.5468$\pm$0.0043 & \textbf{0.2237}$\pm$0.0469 & 0.5243$\pm$0.0176 \\
                          & train & \textbf{0.8341}$\pm$0.0447 & 0.7587$\pm$0.0025 & \textbf{0.8942}$\pm$0.0128 & 0.7615$\pm$0.0134 \\
                          & test  & 0.6360$\pm$0.0715 & \textbf{0.7279}$\pm$0.0371 & 0.6541$\pm$0.0451 & \textbf{0.7189}$\pm$0.0408 \\\midrule
\multirow{3}{*}{PTC\_MR}  & loss  & \textbf{0.3677}$\pm$0.1206 & 0.6722$\pm$0.0066 & \textbf{0.3222}$\pm$0.0814 & 0.5803$\pm$0.0257 \\
                          & train & \textbf{0.8194}$\pm$0.0855 & 0.5948$\pm$0.0188 & \textbf{0.8445}$\pm$0.0637 & 0.6794$\pm$0.0230 \\
                          & test  & 0.4941$\pm$0.0681 & \textbf{0.5882}$\pm$0.0696 & 0.5118$\pm$0.1387 & \textbf{0.5412}$\pm$0.0606 \\ \bottomrule
\end{tabular}
\end{table}

We also present some figures showing the training curves. From the figures, we also clearly observe the phenomenon that respecting lexical invariance results in much more stable training curves and reduces overfitting.

\begin{figure}[h]
    \includegraphics[width=0.33\textwidth]{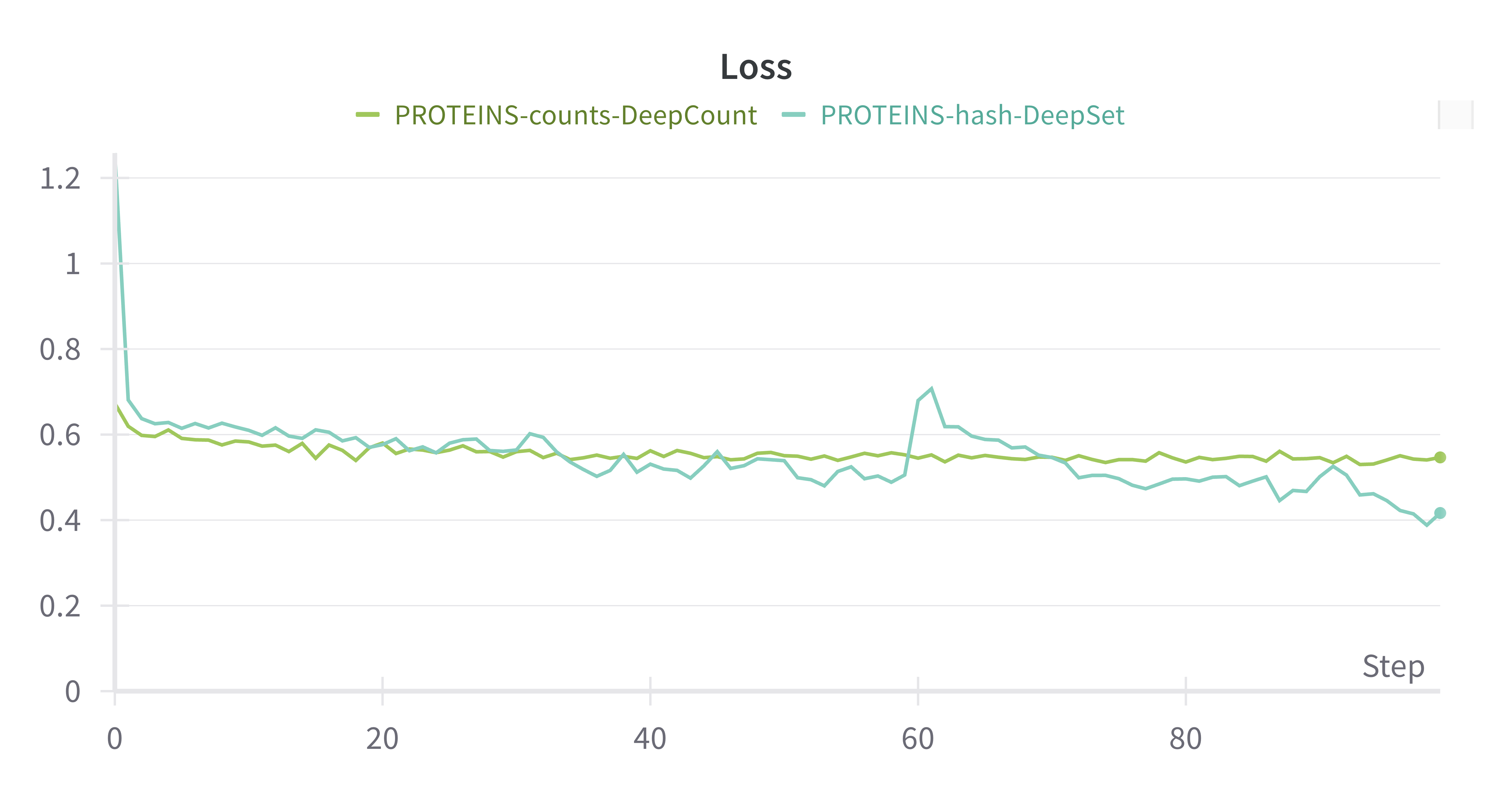}
    \includegraphics[width=0.33\textwidth]{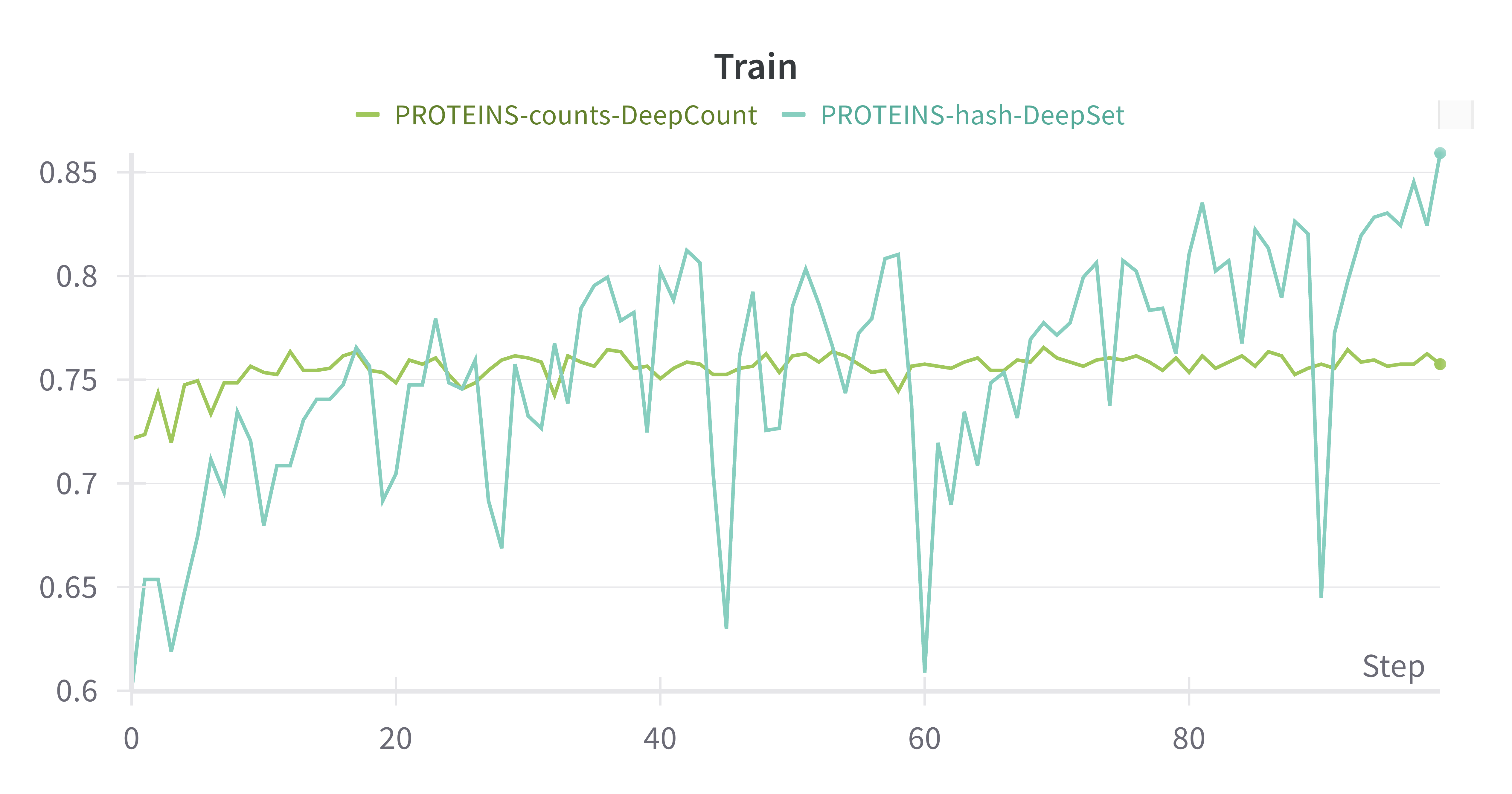}
    \includegraphics[width=0.33\textwidth]{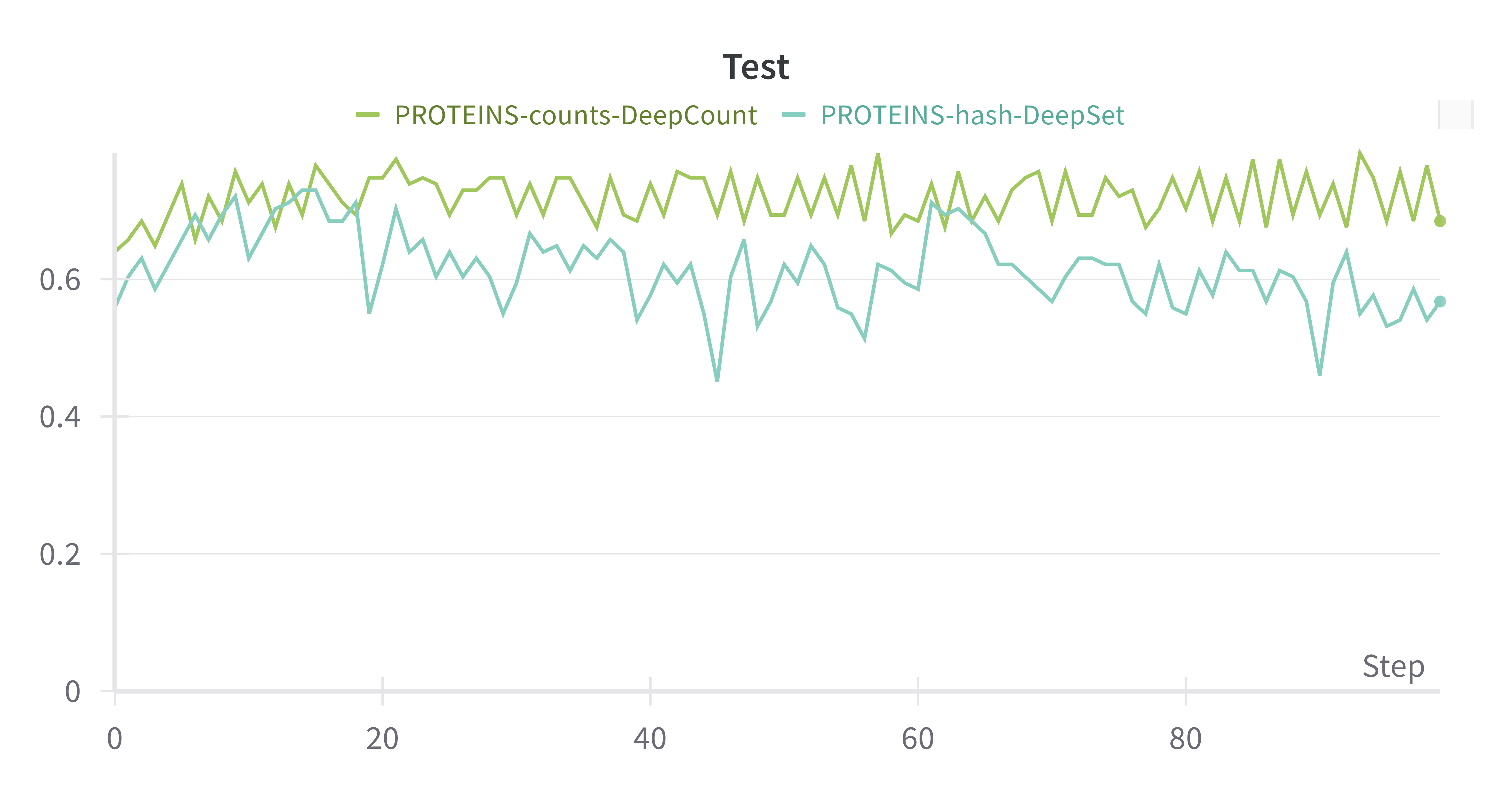}
    \caption{The training curves of DeepSet and DeepCount on PROTEINS.}
    \label{fig:proteins-set}
\end{figure}

\begin{figure}[h]
    \includegraphics[width=0.33\textwidth]{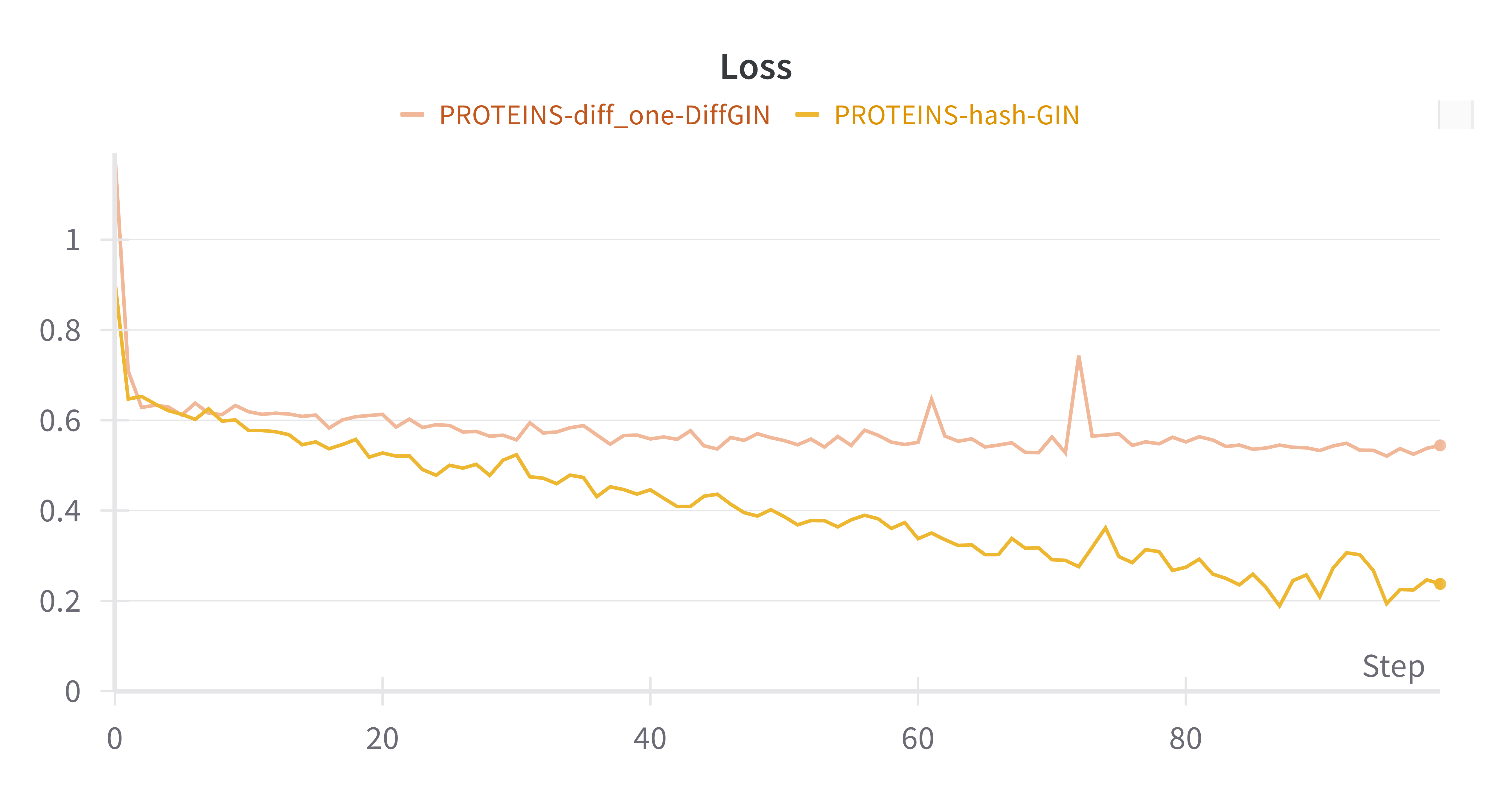}
    \includegraphics[width=0.33\textwidth]{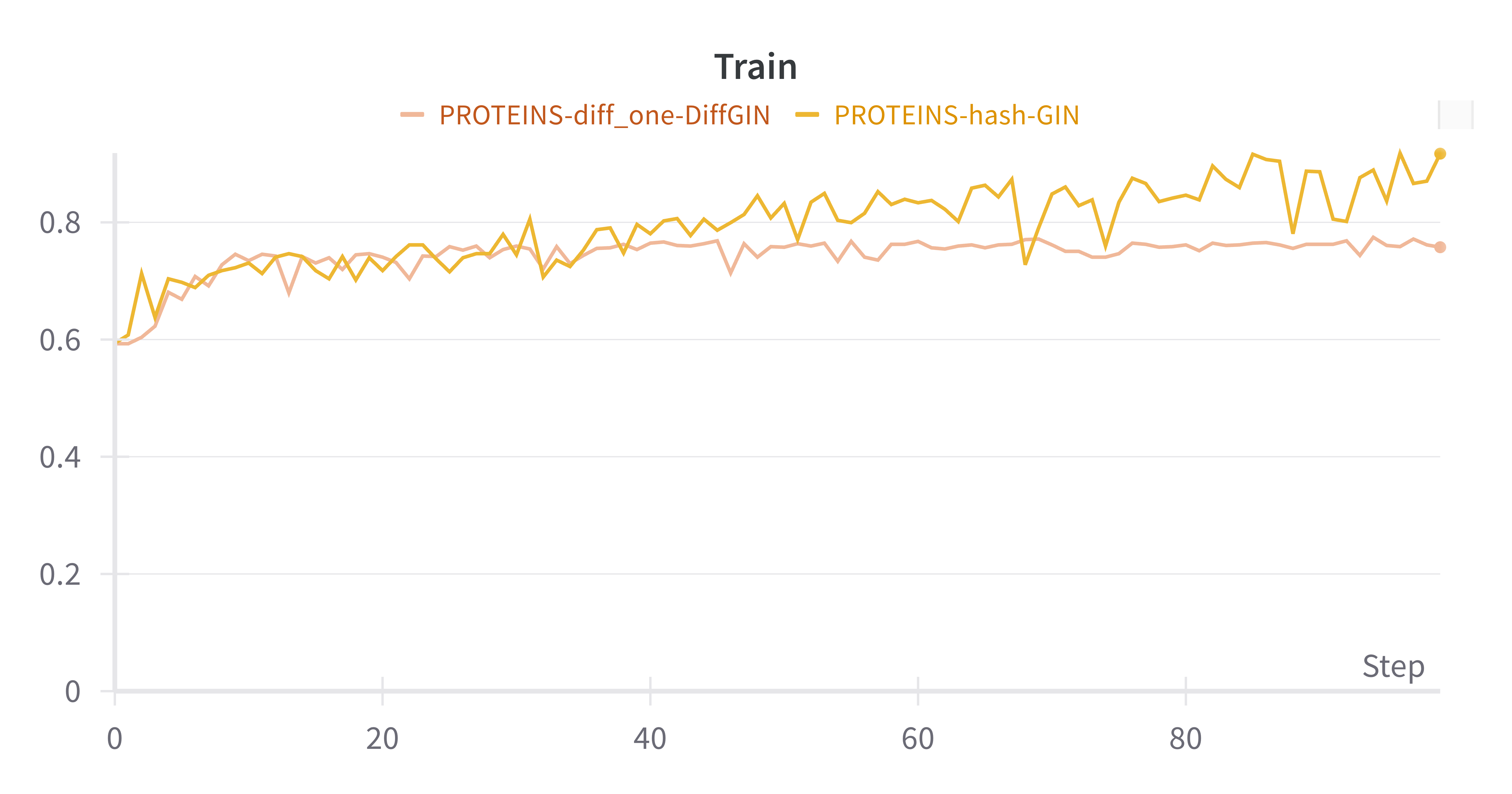}
    \includegraphics[width=0.33\textwidth]{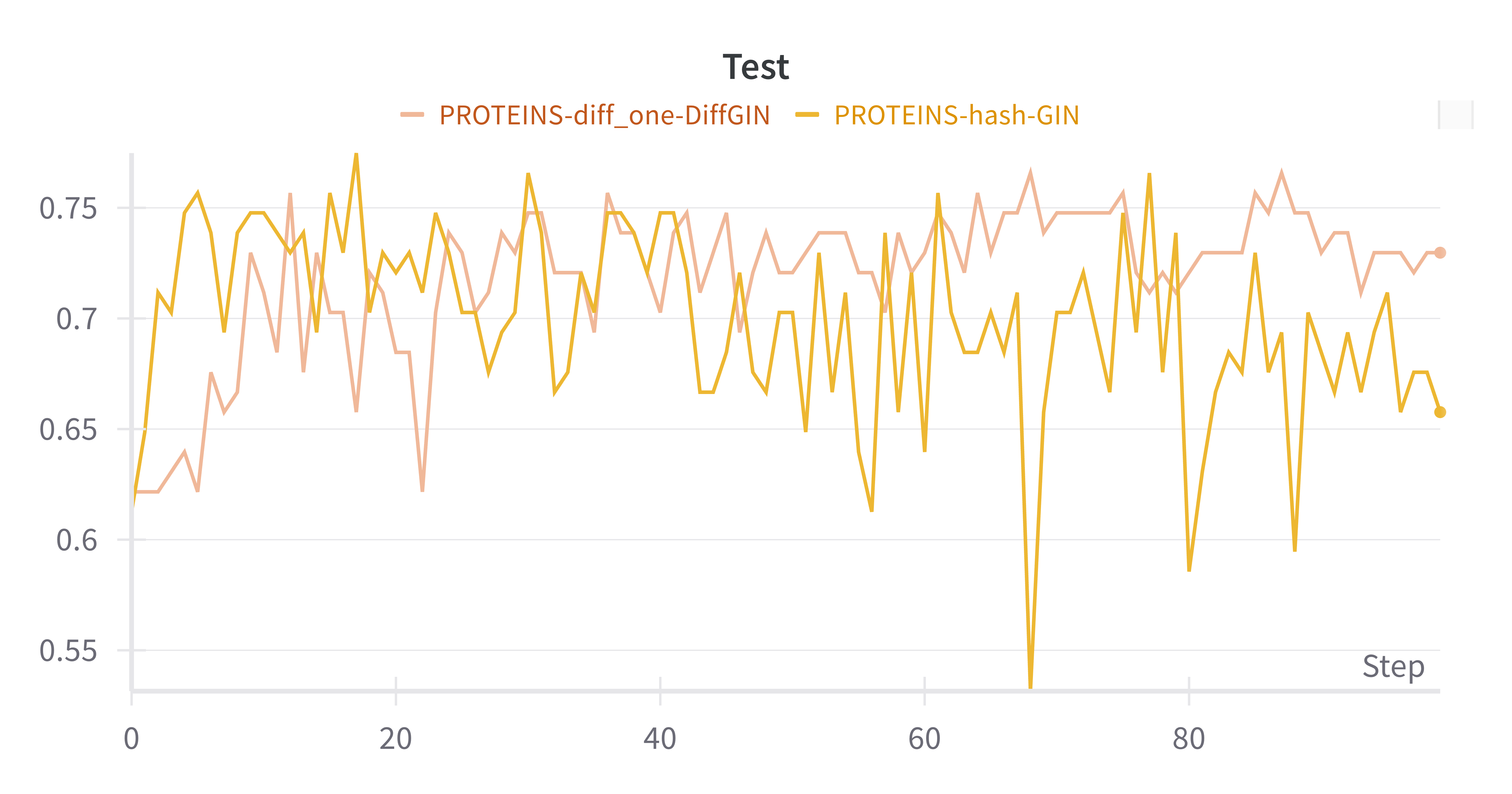}
    \caption{The training curves of GIN and DiffGIN on PROTEINS.}
    \label{fig:proteins-graph}
\end{figure}

\bibliographystyle{unsrt}
\bibliography{references}

\end{document}